# A Circuit-Level Amoeba-Inspired SAT Solver

N. Takeuchi, *Member, IEEE*, M. Aono, Y. Hara-Azumi, and C. L. Ayala, *Member, IEEE*

*Abstract*—AmbSAT (or AmoebaSAT) is a biologically-inspired stochastic local search (SLS) solver to explore solutions to the Boolean satisfiability problem (SAT). AmbSAT updates multiple variables in parallel at every iteration step, and thus AmbSAT can find solutions with a fewer number of iteration steps than some other conventional SLS solvers for a specific set of SAT instances. However, the parallelism of AmbSAT is not compatible with general-purpose microprocessors in that many clock cycles are required to execute each iteration; thus, AmbSAT requires special hardware that can exploit the parallelism of AmbSAT to quickly find solutions. In this paper, we propose a circuit model (hardware-friendly algorithm) that explores solutions to SAT in a similar way to AmbSAT, which we call circuit-level AmbSAT (CL-AmbSAT). We conducted numerical simulation to evaluate the search performance of CL-AmbSAT for a set of randomly generated SAT instances that was designed to estimate the scalability of our approach. Simulation results showed that CL-AmbSAT finds solutions with a fewer iteration number than a powerful SLS solver, ProbSAT, and outperforms even AmbSAT. Since CL-AmbSAT uses simple combinational logic to update variables, CL-AmbSAT can be easily implemented in various hardware.

*Index Terms*—SAT solver, stochastic-local-search solver, bio-inspired computing

## I. Introduction

THE Boolean satisfiability problem (SAT) is a problem to determine if all the given logical constraints, or Boolean formula, can be satisfied and is classified as a nondeterministic polynomial time (NP)-complete problem [1], which indicates that all NP problems, including many practical real-world problems, can be reduced to SAT [2]. Therefore, it is important to develop algorithms and hardware that can find solutions to SAT as fast as possible. So far, many types of algorithms have been proposed: systematic solvers such as Davis-Putnam-Logemann-Loveland (DPLL) [3], Chaff [4], and MiniSAT [5]; and stochastic-local-search (SLS) solvers such as GSAT [6], WSAT [7], and ProbSAT [8]. In general, these algorithms change, or flip, a single variable at each iteration step while searching solutions.

Recently, another type of algorithms has been proposed: AmbSAT (or AmoebaSAT) [9], [10], which changes multiple variables in parallel at each iteration step. AmbSAT is an SLS solver that is inspired by the complex spatiotemporal dynamics of a single-celled amoeba of the true slime mold *Physarum polycephalum*, which deforms into optimal shapes to maximize favorable nutrient absorption and minimize the risk of being exposed to aversive light stimuli [11], [12]. For some randomly generated SAT instances, AmbSAT can find solutions to SAT with a much fewer number of iteration steps than WSAT [7], which is a simple SLS solver, because at each iteration step the former can travel a longer distance in the search space than the latter [10], [13].

However, conventional microprocessors require relatively large computation time to run AmbSAT because the number of clock cycles required to execute an iteration step increases rapidly as the number of variables increases and more variables are flipped in parallel [13]. This indicates that AmbSAT is not compatible with general-purpose microprocessors and should be implemented in special hardware that can exploit the algorithmic parallelism in AmbSAT by using physical parallelism in hardware. Unfortunately, AmbSAT was found to have difficulty in such hardware implementation; Nguyen *et al.* designed a circuit dedicated to AmbSAT using a field-programmable gate array (FPGA) [14], but it turned out that many clock cycles were required for each iteration step due to complex conditional branches defined in AmbSAT, which spoil the physical parallelism in the circuit. Therefore, in order to achieve hardware that can quickly find solutions to SAT by taking advantage of algorithmic and physical parallelisms, circuit-friendly approaches are required, i.e., simple circuit models for AmbSAT should be conceived.

In this paper, we propose a circuit model (hardware-friendly algorithm) that searches for solutions to SAT in a similar manner to AmbSAT, which we call circuit-level AmbSAT (CL-AmbSAT). CL-AmbSAT is circuit-friendly because it uses only simple combinational logic (without complex conditional branches) to update variables. Following a brief explanation on AmbSAT in Section II, we show the details of CL-AmbSAT in Section III, where we introduce two versions of CL-AmbSAT. In Section IV, we show the results of numerical experiments to compare the search performance of CL-AmbSAT with those of

The present study was supported by PRESTO of the Japan Science and Technology Agency (JST) (Grant Nos. JPMJPR1528 and JPMJPR1321) and KAKENHI of the Japan Society for the Promotion of Science (JSPS) (Grant No. 17H04677). *(Corresponding author: Naoki Takeuchi).*

N. Takeuchi is with the Institute of Advanced Sciences, Yokohama National University, 79-5 Tokiwadai, Hodogaya, Yokohama 240-8501, Japan; PRESTO, Japan Science and Technology Agency, 4-1-8 Honcho, Kawaguchi, Saitama 332-0012, Japan (e-mail: takeuchi-naoki-kx@ynu.ac.jp).

M. Aono is with the Faculty of Environment and Information Studies, Keio University, 5322 Endo, Fujisawa, Kanagawa 252-0882, Japan (e-mail: aono@sfc.keio.ac.jp).

Y. Hara-Azumi is with School of Engineering, Tokyo Institute of Technology, S3-50, 2-12-1 Ookayama, Meguro, Tokyo 152-8552, Japan (e-mail: hara@cad.ict.e.titech.ac.jp).

C. L. Ayala is with the Institute of Advanced Sciences, Yokohama National University, 79-5 Tokiwadai, Hodogaya, Yokohama 240-8501, Japan (e-mail: ayala-christopher-pz@ynu.ac.jp).

AmbSAT and ProbSAT [8], which is a powerful SLS solver for randomly generated SAT instances.

It should be noted that, for hardware implementation [15], SAT solvers should be selected depending on applications. For non-embedded system applications, such as software debugging and verification, the manageable problem size is prioritized over the solution time; thus, simple solvers with medium-grained parallelism, such as WSAT, are suitable. Kanazawa *et al.* designed FPGA-dedicated WSAT [16]-[18], which can handle more than $10^5$ variables. On the other hand, for embedded system applications, such as the control of walking robots (the problem size of which is up to hundreds of variables) [19], solvers with fine-grained parallelism are preferred because real-time responses are crucial. Therefore, the hardware implementations dedicated to AmbSAT or CL-AmbSAT are suitable for embedded system applications.

## II. AmbSAT

SAT is a problem to determine if the given logical formula $f$ has an assignment of the variables $x_i \in \{0, 1\}$ that satisfies $f = 1$, where $i \in \{0, 1, …, N\}$, and $N$ is the number of variables. For example, a formula in the conjunctive normal form (CNF) $f_1 = (x_1 \lor x_2 \lor \neg x_3) \land (\neg x_2 \lor x_3 \lor \neg x_4) \land (x_2 \lor x_3 \lor \neg x_4) \land (\neg x_3 \lor x_4 \lor \neg x_1) \land (x_3 \lor x_4 \lor \neg x_1) \land (\neg x_4 \lor x_1 \lor \neg x_2) \land (x_4 \lor x_1 \lor \neg x_2) \land (\neg x_1 \lor x_2 \lor \neg x_3) \land (x_1 \lor x_2 \lor x_3)$ has a unique solution $(x_1, x_2, x_3, x_4) = (1, 1, 1, 1)$. SAT is called 3-SAT when every clause (parenthesis such as $(x_1 \lor x_2 \lor \neg x_3)$) has at most three literals ($x_i$ or $\neg x_i$), as with the above example. It has been proven that 3-SAT is also NP-complete [20]. In this paper, we treat only 3-SAT for simplicity because an arbitrary SAT instance can be reduced to a 3-SAT instance.

There are several versions of AmbSAT; however, an iteration of AmbSAT commonly includes the following three procedures. (1) The logic states of all the variables are observed. (2) The variables are updated in parallel such that the variables that do not satisfy given constraints are flipped and the others are conserved. The variables that cause contradiction, which will be explained later, are also flipped. (3) The update of the variables fails stochastically (i.e., the variables are stochastically flipped, regardless of given constraints), which is required to avoid deadlocked states, where variables keep evolving but never reach a solution [11]. The variables eventually stop changing as procedures 1 through 3 are iterated, which ensures that the assignment of the variables corresponds to a solution, where all the constraints are satisfied [10].

Figure 1 shows an example of the time evolution of parameters while AmbSAT explores solutions to the following SAT instance: $f_2 = (x_1 \lor x_2 \lor x_3) \land (x_1 \lor \neg x_2 \lor x_4) \land (\neg x_1 \lor x_2 \lor x_4) \land (\neg x_3 \lor x_4 \lor x_5)$. $X_{i,v}(t) \in \{-1, 0, 1\}$, where $t$ represents the current iteration step, is a parameter to determine if $x_i(t)$ should be $v \in \{0, 1\}$, as follows:

$$x_i(t) = \begin{cases} 0 & \text{if } X_{i,0}(t) = 1 \text{ and } X_{i,1}(t) \le 0, \\ 1 & \text{if } X_{i,0}(t) \le 0 \text{ and } X_{i,1}(t) = 1, \\ x_i(t-1) & \text{otherwise.} \end{cases} \quad (1)$$

By using a pair of parameters, $X_{i,0}(t)$ and $X_{i,1}(t)$, each variable $x_i(t)$ can take one of the following four states: $x_i(t)$ should be 0 ($X_{i,0} = 1$ and $X_{i,1} \le 0$); $x_i(t)$ should be 1 ($X_{i,0} \le 0$ and $X_{i,1} = 1$); $x_i(t)$ should be kept ($X_{i,0} \le 0$ and $X_{i,1} \le 0$); the logic state of $x_i(t)$ cannot be determined ($X_{i,0} = 1$ and $X_{i,1} = 1$), which we call contradiction. These four states clarify if the variable should be flipped or conserved. All the variables are checked at every iteration step. Then, $X_{i,v}(t)$ decreases when $x_i$ does not satisfy some constraints and/or causes contradiction if $x_i$ becomes $v$; otherwise, $X_{i,v}(t)$ increases. In Fig. 1, $X_{i,v}(t)$'s with blue diagonal lines decrease at the next iteration step $t+1$. In this way, the variables that do not satisfy constraints are flipped, and the others are conserved. Red $X_{i,v}(t)$'s show stochastically changed parameters. As all $X_{i,v}(t)$'s, or $x_i(t)$'s, are updated at every iteration step, the variables finally stabilize at $t = 6$, which indicates that a solution $\mathbf{x} = (x_1, x_2, x_3, x_4, x_5) = (1, 1, 1, 1, 0)$ is found. One of the features of AmbSAT is that the variables can return to an unsatisfied state, where some of the given constraints are not satisfied, after a solution is found; thus, AmbSAT can explore multiple solutions during a time evolution, which is important in some applications such as simulation of chemical reactions [21]. In Fig. 1, the variables return to an unsatisfied state at $t = 9$, but a solution is found again at $t = 10$. More details regarding AmbSAT can be found in the literature [9], [10].

Fig. 1. Time evolution of the parameters $X_{i,v}(t)$ while AmbSAT explores solutions to $f_2$. A solution $\mathbf{x} = (1, 1, 1, 1, 0)$ is found at $t = 6$.

## III. Circuit-Level AmbSAT (CL-AmbSAT)

We show two versions of CL-AmbSAT: version 1 uses minimal circuits but flips variables carelessly; version 2 adds some extra circuits to carefully flip variables.

### A. Version 1: CL-AmbSAT1

In CL-AmbSAT, each variable $x_i(t)$ is represented by a small circuit unit called a variable cell, and the variable cells are interconnected according to the given logical constraints. Figure 2 illustrates a schematic of a variable cell in CL-AmbSAT version 1 (CL-AmbSAT1). The variable cell represents $x_i(t)$ and is composed of basic logic gates, flip-flops (FFs), a majority (MAJ) gate, and stochastic gates (SGs). In the following, we will explain how each circuit component works

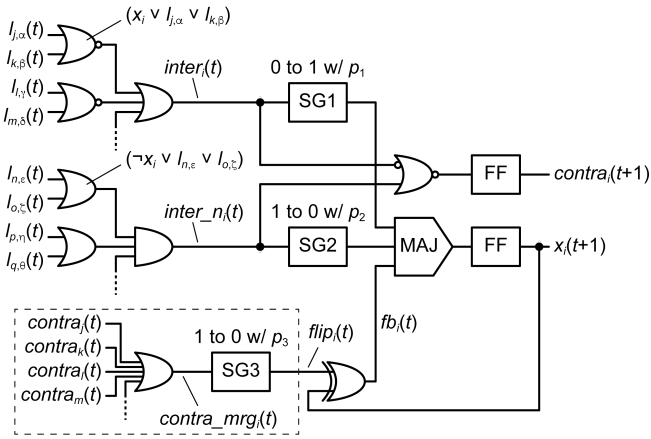

Fig. 2. A variable cell representing $x_i(t)$ in CL-AmbSAT1. $inter_i(t)$ and $inter\_n_i(t)$ change $x_i(i)$ to make true the clauses including $x_i$ and $\neg x_i$. $contra_j(t)$, $contra_k(t)$, and others flip $x_i(i)$ when there is a possibility that $x_i(i)$ causes contradiction in $x_j(t)$, $x_k(t)$, and others.

to explore solutions.

Each clause in the given logical constraints is represented by a NOR or OR gate. The NOR gate that receives $l_{j,\alpha}(t)$ and $l_{k,\beta}(t)$ corresponds to a clause $(x_i \vee l_{j,\alpha} \vee l_{k,\beta})$, where $l_{j,\alpha}$ and $l_{k,\beta}$ are literals that represent $l_{j,\alpha} = x_j$ ($l_{j,\alpha} = \neg x_j$) for $\alpha = 0$ ($\alpha = 1$). When both $l_{j,\alpha}$ and $l_{k,\beta}$ are 0, the NOR gate outputs a logic 1 to change $x_i$ to 1 and make the clause $(x_i \vee l_{j,\alpha} \vee l_{k,\beta})$ true. Similarly, the OR gate that receives $l_{n,\varepsilon}(t)$ and $l_{o,\zeta}(t)$ corresponds to a clause $(\neg x_i \vee l_{n,\varepsilon} \vee l_{o,\zeta})$. When both $l_{n,\varepsilon}$ and $l_{o,\zeta}$ are 0, the OR gate outputs a logic 0 to change $x_i$ to 0 and make the clause $(\neg x_i \vee l_{n,\varepsilon} \vee l_{o,\zeta})$ true. The outputs of the NOR and OR gates are merged into $inter_i(t)$ and $inter\_n_i(t)$, respectively, as follows:

$$inter_i(t) = \neg(l_{j,\alpha} \vee l_{k,\beta}) \vee \neg(l_{l,\gamma} \vee l_{m,\delta}) \vee \cdots, \quad (2)$$

$$inter\_n_i(t) = (l_{n,\varepsilon} \vee l_{o,\zeta}) \wedge (l_{p,\eta} \vee l_{q,\theta}) \wedge \cdots. \quad (3)$$

$inter_i(t)$ becomes 1 if any of the clauses including $x_i$ require $x_i$ to become 1, and $inter\_n_i(t)$ becomes 0 if any of the clauses including $\neg x_i$ require $x_i$ to become 0.

CL-AmbSAT (for both versions 1 and 2) can undergo contradiction when $inter_i(t) = 1$ and $inter\_n_i(t) = 0$: $x_i$ cannot be either 0 or 1. To detect this contradiction, the contradiction-detection signal $contra_i(t+1)$ is calculated using $inter_i(t)$ and $inter\_n_i(t)$, as follows:

$$contra_i(t+1) = \neg(\neg inter_i(t) \vee inter\_n_i(t)). \quad (4)$$

$contra_i(t+1) = 1$ when $inter_i(t) = 1$ and $inter\_n_i(t) = 0$; otherwise, $contra_i(t+1) = 0$. $contra_i(t+1)$ is distributed to the other variable cells that share some clauses with $x_i$ in the given logical constraints. In Fig. 2, assuming $x_i$ shares clauses with $x_i$, $x_j$, $x_k$, $x_l$, and others, the variable cell $x_i(t)$ receives $contra_j(t)$, $contra_k(t)$, $contra_l(t)$, $contra_m(t)$, and others. The contradiction-detection signals are merged into $contra\_mrg_i(t)$ by an OR gate, as follows:

$$contra\_mrg_i(t) = contra_j(t) \vee contra_k(t) \vee \cdots. \quad (5)$$

$contra\_mrg_i(t)$ becomes 1 if any of the variables that share clauses with $x_i$ undergo contradiction.

Stochastic processes are provided by the three SGs: SG1, SG2, and SG3. The output of SG1 is the same as the input with a probability 1-$p_1$ but is forced to be a logic 1 with a probability

$p_1$. The output of SG2 (SG3) is the same as the input with a probability 1-$p_2$ (1-$p_3$) but is forced to be a logic 0 with a probability $p_2$ ($p_3$). We define a following function to express stochastic processes associated with SG1, SG2, and SG3:

$$s_v(x, p) = \begin{cases} v & \text{with a probability } p, \\ x & \text{otherwise.} \end{cases} \quad (6)$$

For instance, the operation of SG3 is expressed using Eq. 6, as follows: SG3 receives $contra\_mrg_i(t)$ and outputs $flip_i(t) = s_0(contra\_mrg_i(t), p_3)$.

Finally, $x_i$ is updated using a MAJ gate, the output of which is determined by the majority vote of inputs:

$$x_i(t+1) = \text{maj}(s_1(inter_i(t), p_1), s_0(inter\_n_i(t), p_2), fb_i(t)), \quad (7)$$

where $fb_i(t) = x_i(t) \veebar flip_i(t)$ and $\text{maj}(a, b, c) = (a \wedge b) \vee (b \wedge c) \vee (c \wedge a)$. Assuming stochastic processes are neglected ($p_1 = p_2 = p_3 = 0$), the MAJ gate works to flip the variables that do not satisfy constraints and conserve the others. For instance, if $inter_i(t) = 0$, $inter\_n_i(t) = 1$, and $contra\_mrg_i(t) = 0$, then $x_i(t+1) = \text{maj}(0, 1, x_i(t)) = x_i(t)$, i.e., $x_i$ keeps its value. On the other hand, if $inter_i(t) = 0$, $inter\_n_i(t) = 0$, and $contra\_mrg_i(t) = 0$, then $x_i(t+1) = \text{maj}(0, 0, x_i(t)) = 0$, i.e., $x_i$ is changed to 0 to make the clauses including $\neg x_i$ true. Furthermore, if $inter_i(t) = 0$, $inter\_n_i(t) = 1$, and $contra\_mrg_i(t) = 1$, then $x_i(t+1) = \text{maj}(0, 1, \neg x_i(t)) = \neg x_i(t)$, i.e., $x_i$ is flipped to solve contradiction.

Here we show how CL-AmbSAT1 solves $f_2 = (x_1 \vee x_2 \vee x_3) \wedge (x_1 \vee \neg x_2 \vee x_4) \wedge (\neg x_1 \vee x_2 \vee x_4) \wedge (\neg x_3 \vee x_4 \vee x_5)$. Since $f_2$ includes five variables, CL-AmbSAT1 includes five variable cells that are interconnected in accordance with $f_2$. For instance, the $x_1(t)$ cell receives $x_2(t)$, $x_3(t)$, $x_4(t)$, $contra_2(t)$, $contra_3(t)$, and $contra_4(t)$ because $x_1$ shares some clauses with $x_2$, $x_3$, and $x_4$. Note that $x_5(t)$ and $contra_5(t)$ are not provided to the $x_1(t)$ cell because $x_1$ and $x_5$ do not share any clauses in $f_2$. Figure 3 shows an example of the time evolution of $x_i(t)$ and $contra_i(t)$ while CL-AmbSAT1 explores solutions to $f_2$. All $x_i(t)$'s and $contra_i(t)$'s are initialized to 0 at $t = 0$. As with AmbSAT, all $x_i(t)$'s are updated at every iteration step. A red $x_i(t)$ represents a stochastically changed $x_i(t)$. At $t = 0$, $x_1(0)$, $x_2(0)$, and $x_3(0)$ undergo contradiction; thus, at $t = 1$, $contra_1(1) = contra_2(1) = contra_3(1) = 1$. At $t = 2$, the variables that receive $contra_i(t) = 1$

| t | 0 | 1 | 2 | 3 | 4 | 5 | 6 | 7 | 8 | 9 | 10 |
|---|---|---|---|---|---|---|---|---|---|---|---|
| $x_1$ | 0 | 0 | 0 | 1 | 1 | 1 | 1 | 0 | 0 | 0 | 0 |
| $x_2$ | 0 | 0 | 0 | 1 | 1 | 1 | 1 | 1 | 1 | 1 | 1 |
| $x_3$ | 0 | 0 | 0 | 1 | 1 | 1 | 1 | 1 | 1 | 1 | 1 |
| $x_4$ | 0 | 0 | 1 | 1 | 1 | 1 | 1 | 1 | 1 | 1 | 1 |
| $x_5$ | 0 | 0 | 0 | 0 | 0 | 0 | 0 | 0 | 0 | 0 | 0 |
| $contra_1$ | 0 | 1 | 1 | 0 | 0 | 0 | 0 | 0 | 0 | 0 | 0 |
| $contra_2$ | 0 | 1 | 1 | 0 | 0 | 0 | 0 | 0 | 0 | 0 | 0 |
| $contra_3$ | 0 | 1 | 1 | 0 | 0 | 0 | 0 | 0 | 0 | 0 | 0 |
| $contra_4$ | 0 | 0 | 0 | 0 | 0 | 0 | 0 | 0 | 0 | 0 | 0 |
| $contra_5$ | 0 | 0 | 0 | 0 | 0 | 0 | 0 | 0 | 0 | 0 | 0 |

Satisfied

Fig. 3. Time evolution of the variables $x_i(t)$ and parameters $contra_i(t)$ while CL-AmbSAT1 explores solutions to $f_2$. Red $x_i(t)$ shows that $x_i(t)$ is stochastically flipped by SGs. A solution $\mathbf{x} = (1, 1, 1, 1, 0)$ is found at $t = 3$, and another solution $\mathbf{x} = (0, 1, 1, 1, 0)$ is found at $t = 7$.

are stochastically flipped; in Fig. 3, $x_4$ is flipped, but the others fail to be flipped due to stochastic processes provided by SG3. At $t = 3$, the variables stabilize, which indicates that a solution $\boldsymbol{x} = (1, 1, 1, 1, 0)$ is found. As with AmbSAT, CL-AmbSAT1 can find multiple solutions during a time evolution. In Fig. 3, another solution $\boldsymbol{x} = (0, 1, 1, 1, 0)$ is found at $t = 7$ because of stochastic processes provided by SG1 and SG2.

In SLS solvers, probability distributions affect search performances significantly [8]. Therefore, we optimize the probability distributions in CL-AmbSAT1. Figure 4 shows the simulation results of the search performance of CL-AmbSAT1 vs. probability distributions $p_1$, $p_2$, and $p_3$, where Python scripts were used to simulate CL-AmbSAT1. In the simulation, a benchmark instance "uf50-01.cnf", which is a randomly generated 3-SAT instance including 50 variables and 218 clauses, from an online instance library SATLIB [22] was used. Figure 4(a) shows average iteration numbers to find a solution as a function of $p_3$, where $p_1 = p_2 = 0.01$. Average iteration numbers rapidly increase as $p_3$ goes below 0.8, which implies that $p_3$ should be set to a value close to 1. This is because, if $p_3$ is small, too many variables can be flipped by $contra\_mrg_i(t)$ at a time, which lets an assignment of the variables travel a very long distance in the search space even though the assignment of the variables is close to solutions. Therefore, we set $p_3$ to 0.9 in this study. Figure 4(b) shows average iteration numbers to find a solution as a function of $p_1$ and $p_2$, where $p_3 = 0.9$ and we assume $p_1 = p_2$ for simplicity; we treat 0s and 1s equally, so that it is reasonable to assume that the probability of flipping a 0 to a 1 ($p_1$) is the same as that of flipping a 1 to a 0 ($p_2$). It is also possible to set $p_1$ and $p_2$ to different values via a small extension (e.g., the design of the SGs). Average iteration numbers decrease as $p_1$ and $p_2$ decrease, which implies that $p_1$ and $p_2$ should be set to values close to 0. In this study, we set $p_1$ and $p_2$ to be $1/2N$, where $N$ is the number of variables, such that stochastic processes via SG1 and SG2 do not often appear.

### B. Version 2: CL-AmbSAT2

CL-AmbSAT1 uses simple circuit schematics but changes variables carelessly. $flip_i(t) = s_0(contra\_mrg_i(t), p_3)$ flips $x_i(t)$ stochastically to solve contradiction; however, the flip of $x_i(t)$ could deteriorate, rather than improve, the assignment of the variables. Furthermore, $x_i(t)$ is not restored unless $x_i(t)$ is flipped again by $flip_i(t)$. In CL-AmbSAT version 2 (CL-AmbSAT2), $inter_i(t)$, $inter\_n_i(t)$, $contra_i(t)$, and $contra\_mrg_i(t)$ are calculated in the same way as in CL-AmbSAT1, but $flip_i(i)$ is differently calculated to carefully flip variables. In CL-AmbSAT2, the part surrounded by the dashed lines in a variable cell (Fig. 2) is replaced with a circuit shown in Fig. 5. $attm_i(t)$ and $attm\_contra_i(t)$ attempt to change the variables that satisfy $inter_i(t) = 0$ and $inter\_n_i(t) = 1$. The difference between $attm_i(t)$ and $attm\_contra_i(t)$ is that the latter works for the variables that could cause contradiction ($contra\_mrg_i(t) = 1$) and the former works for the others. When $attm_i(t)$ and/or $attm\_contra_i(t)$ is 1, $flip_i(t)$ becomes 1 and flips $x_i(t)$. At the next iteration $t+1$, $attm\_mrg_i(t+1)$ becomes 1, and $flip_i(t+1)$ is forced to be 0 so that $x_i$ is not flipped again. At $t+2$ ($judge(t+2) = 1$), it is determined if $x_i$ is kept or restored, depending on $contra\_mrg_i(t+2)$. If $contra\_mrg_i(t+2) = 0$, the flip of $x_i$ at the iteration $t$ is considered as a success because the flip of $x_i$ might have solved some contradiction; thus, $restore_i(t+2) = 0$, i.e., $x_i$ is kept. If $contra\_mrg_i(t+2) = 1$, the flip of $x_i$ at $t$ is considered as a failure; thus, $restore_i(t+2) = 1$ and $flip_i(t+2) = 1$. i.e., $x_i$ is flipped again and restored to the original value. In this way, CL-AmbSAT2 flips $x_i$ more carefully than CL-AmbSAT1.

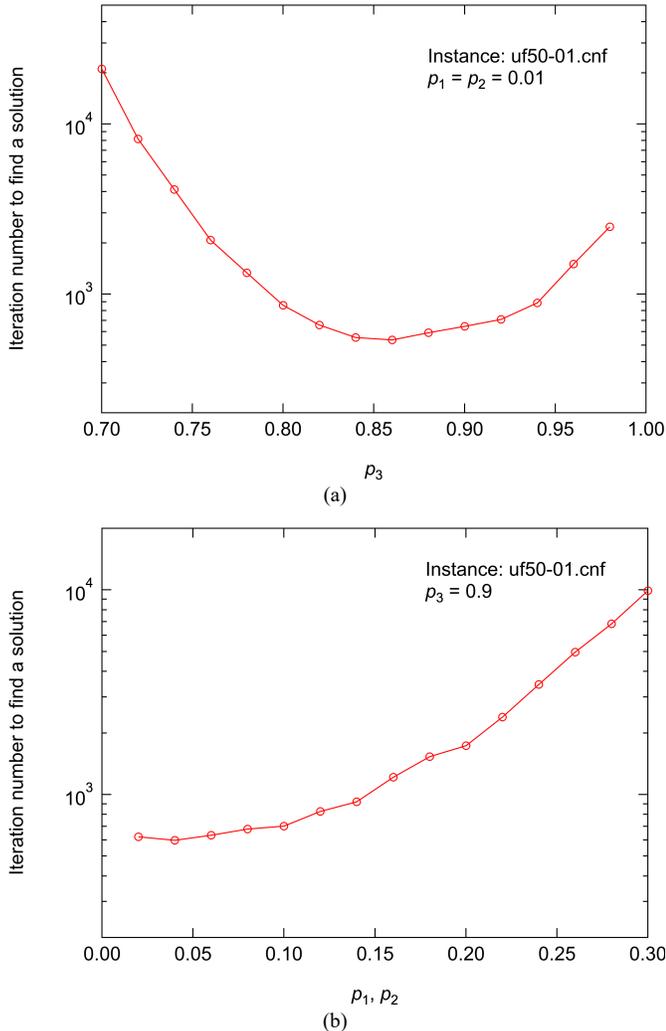

Fig. 4. Search performance vs. probability distributions in CL-AmbSAT2. (a) Iteration number to find a solution to uf50-01 as a function of $p_3$. (b) Iteration number to find a solution to uf50-01 as a function of $p_1$ and $p_2$, where $p_1 = p_2$ for simplicity. The above iteration numbers are averages over 500 trials.

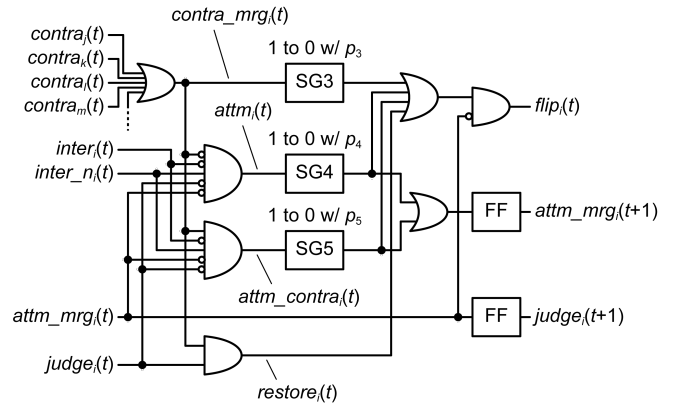

Fig. 5. A part of a variable cell in CL-AmbSAT2, which calculates $flip_i(i)$ in a different way from CL-AmbSAT1. $attm_i(t)$ and $attm\_contra_i(t)$ flip $x_i$ stochastically by letting $flip_i(t)$ be 1; however, $x_i$ returns to the original value if $restore_i(t)$ judges that the flip of $x_i$ is a failure.

Stochastic processes are provided by SG3, SG4, and SG5. We investigated the search performance of CL-AmbSAT2 for different probability distributions in the similar way to Fig. 4, so that we determined that $p_3 = p_4 = 0.95$ and $p_5 = 0.2$. Note that $p_1 = p_2 = 0$ in CL-AmbSAT2.

## IV. NUMERICAL EXPERIMENTS

We investigate the search performance of the two versions of CL-AmbSAT using numerical simulation. For the simulation, we create SAT instances systematically using a benchmark instance "uf50-0100.cnf" from SATLIB, which is a randomly generated 3-SAT instance that includes 50 variables and 218 clauses and has a unique solution. We make a "polymer" instance by connecting $h$ sub-instances of uf50-0100.cnf [13], so that the whole instance has $N = 50 \times h$ variables and $M = 218 \times h$ clauses, where each sub-instance uses different variable names. In this way, we can create an arbitrarily large SAT instance that always has a unique solution, which helps to clarify the scalability of our approach, i.e., extrapolation of the iteration number to find the solution as a function of the problem size $N$. Figure 6 shows the simulation results of average iteration numbers to find a solution as a function of the problem size $N$, where Python scripts were used to simulate CL-AmbSAT. For comparison, simulation results for AmbSAT and ProbSAT [8], which is an SLS solver that won the category of "Core Solvers, Sequential, Random SAT" at SAT Competition 2013, are also shown [13]. The fitting curve for ProbSAT is proportional to $N$, which is reasonable because the $h$ sub-instances in the polymer instance of uf50-0100.cnf are independent of each other. On the other hand, the fitting curves for AmbSAT and CL-AmbSAT are proportional to $\ln N$, which indicates that CL-AmbSAT, as well as AmbSAT, exploits algorithmic parallelism to find solutions, and the benefit of parallelism increases as the problem size increases. For all the problem sizes, CL-AmbSAT2 can find solutions with the fewest average iteration number. One possible reason why CL-AmbSAT is even faster than AmbSAT is that CL-AmbSAT uses multiple SGs to change probability distributions more flexibly. Although only the polymer instance of uf50-0100.cnf was treated in this study, in future, we will treat different instances to more comprehensively understand the search performance of CL-AmbSAT (e.g., what kind of problems CL-AmbSAT can solve more quickly than conventional solvers).

Since in this study we do not specify the hardware that implements CL-AmbSAT, in Fig. 6 we evaluated its performance using iteration numbers, rather than clock cycles or execution time. Importantly, CL-AmbSAT is so circuit-friendly that FPGAs can perform an iteration of CL-AmbSAT in a single clock cycle [23]. On the other hand, ProbSAT and AmbSAT may require more clock cycles to execute each iteration. Therefore, the advantage of CL-AmbSAT would be further extended in hardware implementation.

The implementation of CL-AmbSAT1 using an FPGA is reported in [23]. We compared the hardware complexity of an FPGA-dedicated CL-AmbSAT1 with that of an FPGA-dedicated WSAT [16]. We found that the number of slices for the former was much less than that for the latter; e.g., for the instance uf225-0118.cnf (which includes 225 variables), the slice count for CL-AmbSAT1 was 3,800 on Virtex2, whereas it was 17,234 for WSAT on the same device. We also confirmed that the slice count in the FPGA-dedicated CL-AmbSAT1 increases in proportion to the number of variables. Additionally, we found that each iteration step in CL-AmbSAT1 requires only one clock cycle, and that the clock cycle to find a solution increases logarithmically with regard to the number of variables (although the clock frequency gradually decreases as the number of variables increases).

## V. CONCLUSION

We proposed CL-AmbSAT, which is a circuit mode that finds solutions to SAT in the manner of AmbSAT. In CL-AmbSAT, multiple variables are updated at every iteration step, and SAT solutions are found when the variables stabilize. We optimized the probability distributions in CL-AmbSAT. Numerical experiments showed that, for the polymer instances of uf50-0100.cnf, the number of iteration steps required to find a solution in CL-AmbSAT is proportional to $\ln N$, and that CL-AmbSAT can find solutions with a fewer iteration number than ProbSAT and AmbSAT. Most importantly, CL-AmbSAT updates variables by only using simple combinational logic (unlike AmbSAT, which includes complex conditional branches). Therefore, CL-AmbSAT can be easily implemented in various hardware, including conventional application specific integrated circuits (ASICs) and FPGAs; and even post complementary metal–oxide–semiconductor (post-CMOS) devices.

ACKNOWLEDGMENT

The authors would like to thank S. Kasai for useful discussion.

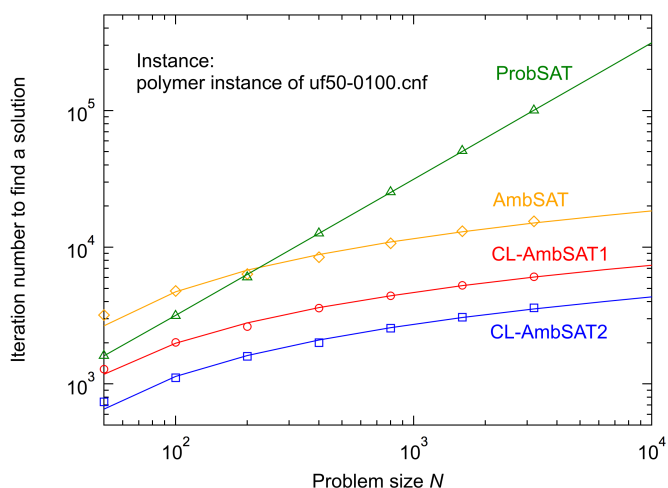

Fig. 6. Comparison of average iteration numbers to find solutions to polymer instances of uf50-0100.cnf among several SAT solvers. Iteration numbers in AmbSAT and CL-AmbSAT increase logarithmically as a function of the problem size. The above iteration numbers are averages over 500 trials.